\documentclass[10pt,twocolumn,letterpaper]{article}

\usepackage[pagenumbers]{cvpr} 

\usepackage{overpic}
\definecolor{cvprblue}{rgb}{0.21,0.49,0.74}
\usepackage[pagebackref,breaklinks,colorlinks,allcolors=cvprblue]{hyperref}

\def\paperID{16261} 
\def\confName{CVPR}
\def\confYear{2025}

\title{One Shot, One Talk: Whole-body Talking Avatar from a Single Image}

\author{Jun Xiang$^{1,2}$ \qquad Yudong Guo$^{1*}$ \qquad Leipeng Hu$^{1}$ \qquad Boyang Guo$^{1}$ \\ Yancheng Yuan$^{2}$ \qquad Juyong Zhang$^{1}$\\
{\tt\small $^{1}$University of Science and Technology of China \qquad $^{2}$The Hong Kong Polytechnic University}
}

\begin{document}

\twocolumn[{%
\renewcommand\twocolumn[1][]{#1}%
\maketitle
\includegraphics[page=1,width=\textwidth]{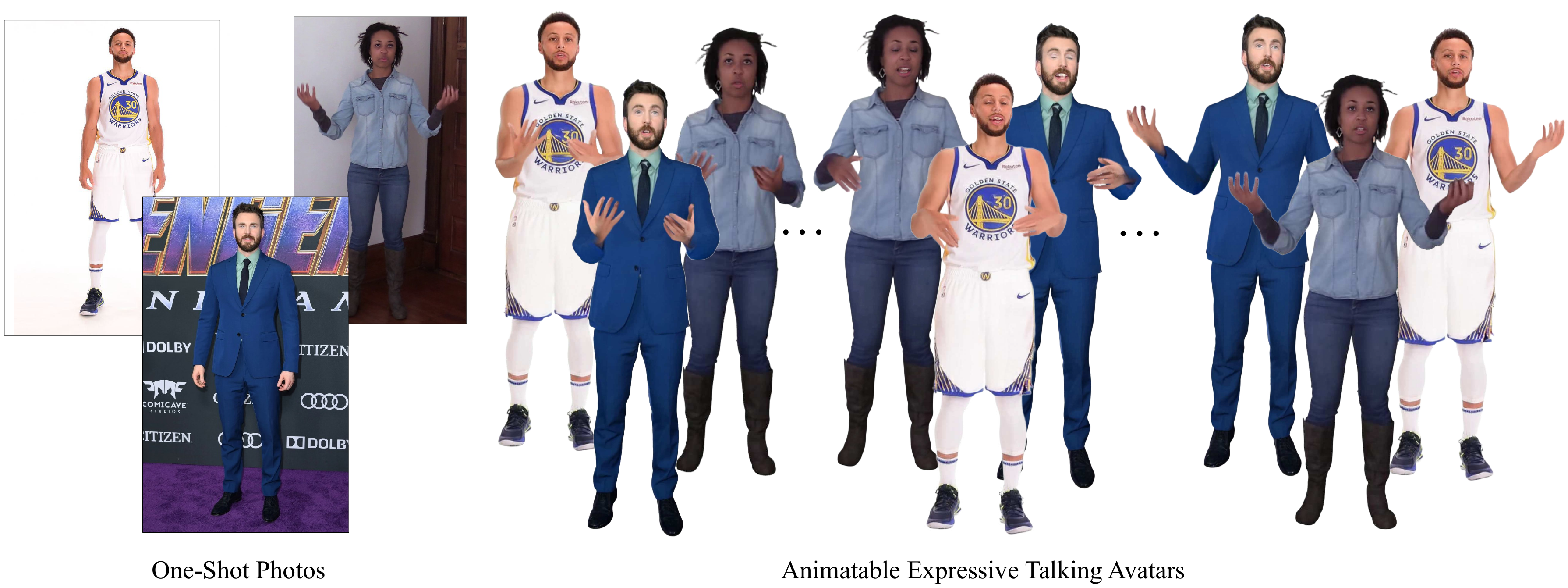}
\vspace{-5mm}
\captionof{figure}{Given a one-shot image (e.g., your favorite photo) as input, our method reconstructs a fully expressive whole-body talking avatar that captures personalized details and supports realistic animation, including vivid body gestures and natural expression changes. Project page: \href{https://ustc3dv.github.io/OneShotOneTalk/}{https://ustc3dv.github.io/OneShotOneTalk/}}
\label{fig:teaser}
\vspace{5mm}
}]

\let\thefootnote\relax\footnotetext{
\vskip -1em
\noindent
{$^{*}$Corresponding author: Yudong Guo.}}

\begin{abstract}
Building realistic and animatable avatars still requires minutes of multi-view or monocular self-rotating videos, and most methods lack precise control over gestures and expressions. To push this boundary, we address the challenge of constructing a whole-body talking avatar from a single image. We propose a novel pipeline that tackles two critical issues: 1) complex dynamic modeling and 2) generalization to novel gestures and expressions. To achieve seamless generalization, we leverage recent pose-guided image-to-video diffusion models to generate imperfect video frames as pseudo-labels. To overcome the dynamic modeling challenge posed by inconsistent and noisy pseudo-videos, we introduce a tightly coupled 3DGS-mesh hybrid avatar representation and apply several key regularizations to mitigate inconsistencies caused by imperfect labels. Extensive experiments on diverse subjects demonstrate that our method enables the creation of a photorealistic, precisely animatable, and expressive whole-body talking avatar from just a single image.
\end{abstract}

\section{Introduction}
\label{sec:intro}
Realistic rendering and precise control of gestures and expressions for whole-body talking avatars hold significant potential for AR/VR applications, such as telepresence and immersive remote conferencing. While extensively studied in the fields of 3D Vision and Graphics, most production-level body avatars still require a light stage~\cite{guo2019relightables, xiang2023drivable} to capture hours or minutes of multi-view frames, making them unsuitable for typical consumer usage. To address this, we focus on the challenging task of constructing an animatable whole-body avatar that supports both photo-realistic rendering and precise 3D control of gestures and expressions, from only a single image. This one-shot pipeline primarily faces the following challenges.

\textbf{Complex dynamic modeling.} Humans exhibit complex gestures and facial movements when communicating with each other. To model the motion space of the entire body, SMPL-X~\cite{SMPL-X:2019} integrates several previous models~\cite{smpl2015,li2017learning,mano2017} to capture body, hand, and facial movements. With the help of these parametric geometric models, recent personalized human avatars learn to model dynamic geometry and appearance by integrating neural textures~\cite{liu2021neural,habermann2021real}, neural radiance fields~\cite{zhao2022humannerf,peng2021neural,jiang2022neuman, weng2022humannerf}, and 3D Gaussians~\cite{lei2024gart,kocabas2024hugs,hu2024gauhuman,hu2024gaussianavatar,qian20243dgs} for dynamic photo-realistic rendering. To capture the full appearance of a whole-body avatar, these methods require complete observations, typically relying on dense input data for supervision, such as multi-view videos or self-rotating monocular videos. Moreover, these approaches primarily focus on body motion and novel view synthesis, falling short in accurately capturing and animating diverse facial expressions and hand movements, which limits their practical applicability. Due to the reliance on dense inputs, realistic and expressive whole-body modeling from a single image remains an unsolved challenge in this field.

\textbf{Generalization to novel gestures and expressions.} Another key challenge lies in the limited ability to generalize to a wide variety of gestures and facial expressions, especially those that deviate significantly from the poses seen during training. This limitation arises from the data-driven nature of current methods, where the training data typically encompasses only a finite set of gestures and expressions. Consequently, when animating avatars, regions or motions that were underrepresented in the training data—such as dynamic clothing, inner mouth movements, or intricate hand gestures—are often poorly synthesized or omitted entirely. These issues are further compounded when working with a single image input, as the limited visual cues often fail to capture the full range of motion and expression. Recently, several approaches~\cite{mimicmotion2024, hu2024animate, xu2024magicanimate, zhu2024champ} have leveraged image and video diffusion models~\cite{sd2022, blattmann2023stable} to achieve image-to-video generation guided by whole-body landmarks~\cite{yang2023effective}. Although various model architectures and training techniques have been explored to improve generation quality, temporal consistency remains a challenge, and these methods have several key limitations. For example, body landmarks fail to disentangle identity and pose, leading to poor cross-identity animation results, with facial and body distortions and identity mismatches. Moreover, the sparse 2D landmarks are insufficient for precisely controlling gestures and facial movements, as they do not accurately capture the underlying dynamic geometry.

In this work, we propose a novel pipeline to address both challenges in the one-shot setting. To generalize to diverse gestures and facial movements, we preprocess the large-scale TED Gesture Dataset~\cite{yoon2019robots} to build a comprehensive whole-body motion space for people talking. We then use these motion sequences as guidance to drive the input single image with a pre-trained whole-body video diffusion model~\cite{mimicmotion2024} and a 3D face animation model~\cite{deng2024}. This enables us to generate various video sequences of the target person performing different gestures and expressions. However, directly using these pseudo labels to train a body avatar leads to unsatisfactory results. First, as noted above, current diffusion-based body animation methods struggle with temporal consistency, identity preservation, and motion alignment. Directly employing them as labels results in significant blurring, distortion, and identity degradation. Furthermore, as we adapt two individual approaches for generating pseudo labels for body gestures and facial expressions, respectively. Due to inconsistencies in camera spaces and rendering procedures, merging them into a unified avatar representation poses additional challenges.

To tackle the challenge of complex dynamic modeling, we utilize both the single input image and imperfect pseudo labels to train a hybrid mesh-3DGS avatar representation, constrained by several carefully designed regularizations. The single input image provides an accurate, though incomplete, appearance for the body avatar, while the pseudo videos offer imperfect but more complete visual cues. For the pseudo video labels, instead of using per-pixel losses, we employ a perceptual-based loss term~\cite{zhang2018unreasonable}, which helps achieve reasonable appearance modeling while alleviating misalignment in the pseudo labels. 

To further alleviate the inconsistencies caused by the pseudo labels, we adopt a tightly coupled mesh-3DGS hybrid avatar representation. By introducing Laplacian smoothing and normal-consistency regularization on the deformed body mesh, we ensure that the structure of the 3D Gaussians used for rendering is well-constrained. Finally, we supervise the avatar representation using both gesture and head video pseudo labels, enabling the creation of a photorealistic, precisely animatable, and expressive whole-body avatar.

In summary, our contributions include the following aspects:
\begin{itemize}
\item We introduce a novel pipeline that overcomes the key challenges of building a whole-body expressive talking avatar from a single image.
\item Our diffusion guidance strategy effectively extracts valuable knowledge from imperfect diffusion outputs and combines it with the limited information from the input image, enabling complete modeling of the talking avatar. 
\item Our carefully designed 3DGS-mesh coupled avatar representation, along with essential regularization techniques, facilitates accurate modeling of diverse subjects and stabilizes the optimization process.
\end{itemize}

\section{Related Work}
\label{sec:relate}

\noindent\textbf{Human Gaussian Splatting.} 3D Gaussian Splatting~\cite{kerbl20233d} is the state-of-the-art method for scene reconstruction and novel view synthesis (NVS), offering superior rendering speed and visual quality. This has significantly influenced human avatar studies. Methods using multi-view videos~\cite{Pang_2024_CVPR, zielonka2023drivable, moreau2024human, qian20243dgs, jiang2024hifi4g} demonstrate excellent performance, with unique designs such as ASH~\cite{Pang_2024_CVPR}, which achieves efficient Gaussian learning via mesh UV parameterization.
For monocular video input, most human Gaussian splatting methods~\cite{hu2024gauhuman, hu2024gaussianavatar, shao2024splattingavatar, lei2024gart, hu2024expressive, moon2024exavatar} link Gaussian fields to parametric mesh models, often using additional regularization terms. ExAvatar~\cite{moon2024exavatar} applies connectivity-based regularizers to short, casually captured videos.
For sparse-view images, GPS-Gaussian~\cite{zheng2024gps} achieves real-time human NVS by encoding human priors into the network, while HumanSplat~\cite{pan2024humansplat} generates high-quality static reconstructions from one-shot inputs.
In contrast, our method is the first human Gaussian approach capable of recovering a realistic, animatable talking avatar from just a single image.

\noindent\textbf{Avatar Reconstruction from Few-Shot Images.} Some works~\cite{dong2023ag3d, jiang2023humangen, xiong2023get3dhuman} apply 3D GAN inversion for one-shot human reconstruction, but they struggle with preserving personal details and generalization. PIFu~\cite{saito2019pifu} and subsequent works~\cite{saito2020pifuhd, xiu2022icon, xiu2023econ, corona2023structured, zheng2021pamir} introduce pixel-aligned features and neural fields for image-based human reconstruction. An alternative approach leverages diffusion priors to fill in missing details, such as training human-centered diffusion models~\cite{pan2024humansplat, ho2024sith, albahar2023single}, using novel-view diffusion results for additional supervision~\cite{liu2024human, li2024pshuman, Zhang_2024_CVPR}, and employing Score Distillation Sampling (SDS)~\cite{poole2022dreamfusion} to generate 3D avatars from 2D priors~\cite{huang2024tech, yang2024have, zhang2024humanref, xiu2024puzzleavatar}. However, these methods focus on static scenes and overlook dynamic human motion, limiting their ability to capture human dynamics. ELICIT~\cite{huang2022elicit} uses CLIP~\cite{radford2021learning} for semantic understanding, but it fails to handle hand and facial motions, restricting expressive animation capabilities. In contrast, our approach leverages priors from a pose-guided human video diffusion model, capturing both human appearance and dynamics, and enabling expressive full-body animations, particularly in the hands and face.

\noindent\textbf{Pose-Guided Human Video Diffusion.} Pose-guided human video diffusion models~\cite{hu2024animate, mimicmotion2024, zhu2024champ, xu2024magicanimate, feng2023dreamoving} directly generate animated videos from a reference image and pose sequence, bypassing traditional 3D reconstruction and rendering processes. The success of these models depends on the quality of training data, model design, and pose guidance. Some approaches~\cite{hu2024animate, feng2023dreamoving, xu2024magicanimate, zhu2024champ} incorporate temporal layers to ensure smooth transitions, inspired by AnimateDiff~\cite{guo2023animatediff}, while others~\cite{mimicmotion2024} use video diffusion models~\cite{blattmann2023stable} for dynamic sequences.
Pose guidance is typically provided by OpenPose~\cite{hu2024animate, mimicmotion2024}, DensePose~\cite{xu2024magicanimate}, depth maps~\cite{feng2023dreamoving}, or SMPL~\cite{zhu2024champ}. MimicMotion~\cite{mimicmotion2024} improves pose accuracy with a confidence-aware strategy, and Make-Your-Anchor~\cite{huang2024make} personalizes outputs by fine-tuning models on identity-specific images. We adopt MimicMotion for its superior performance in handling hand regions.
Despite their strengths, these 2D models still face challenges, such as image distortion, identity changes, and pose misalignment, due to the lack of 3D understanding. To address these, we leverage optimized avatar representations with carefully designed constraints, improving consistency and naturalness in the generated animations.

\section{Method}
\label{sec:method}

\begin{figure*}[htb]
  \centering
  \includegraphics[width=1\linewidth]{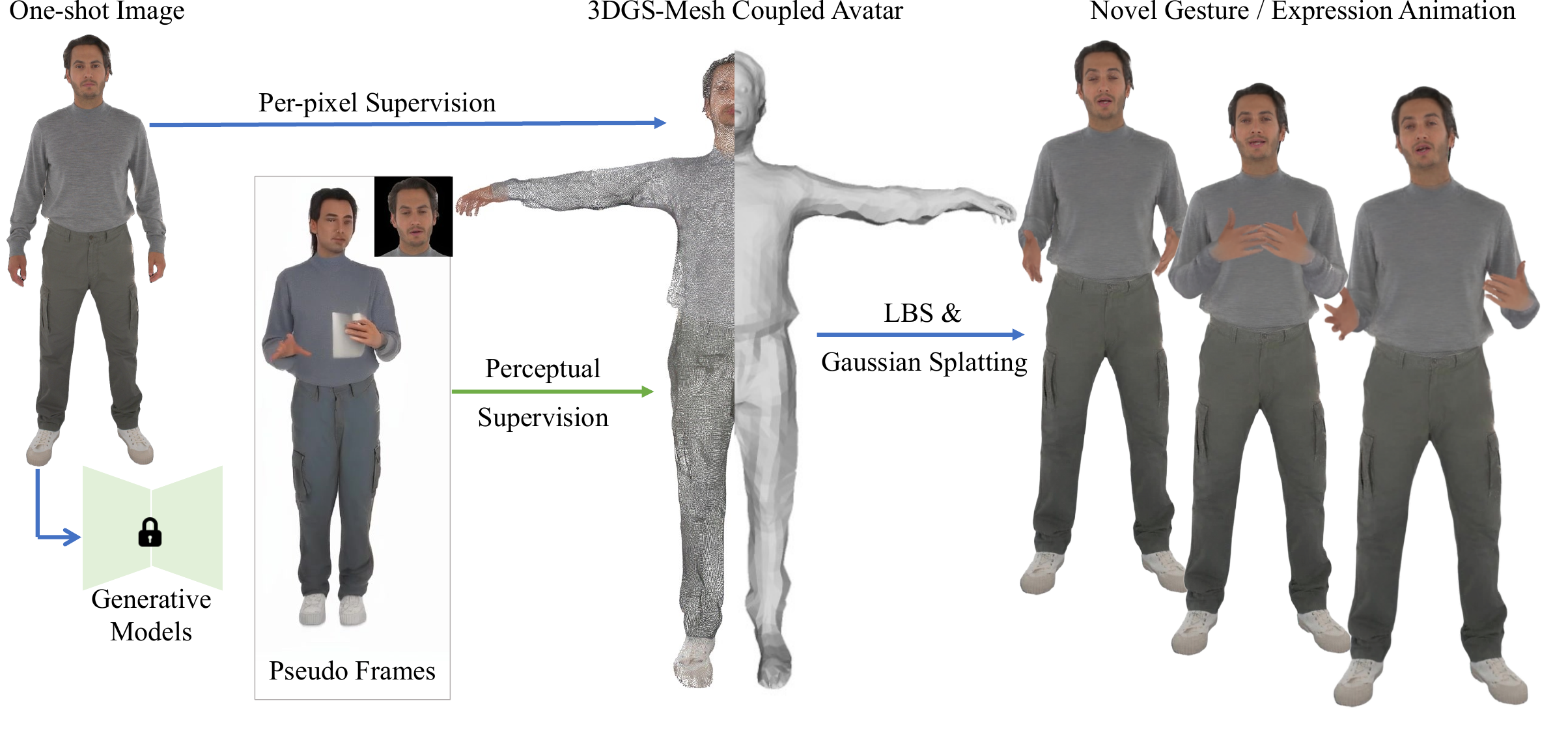}
  \caption{Overview. Our method constructs an expressive whole-body talking avatar from a single image. We begin by generating pseudo body and head frames using pre-trained generative models, driven by a collected video dataset with diverse poses. Per-pixel supervision on the input image, perceptual supervision on imperfect pseudo labels, and mesh-related constraints are then applied to guide the 3DGS-mesh coupled avatar representation, ensuring realistic and expressive avatar reconstruction and animation.}
  \label{fig:pipeline}
\end{figure*}

\begin{figure*}
  \centering
  \includegraphics[width=1.0\linewidth]{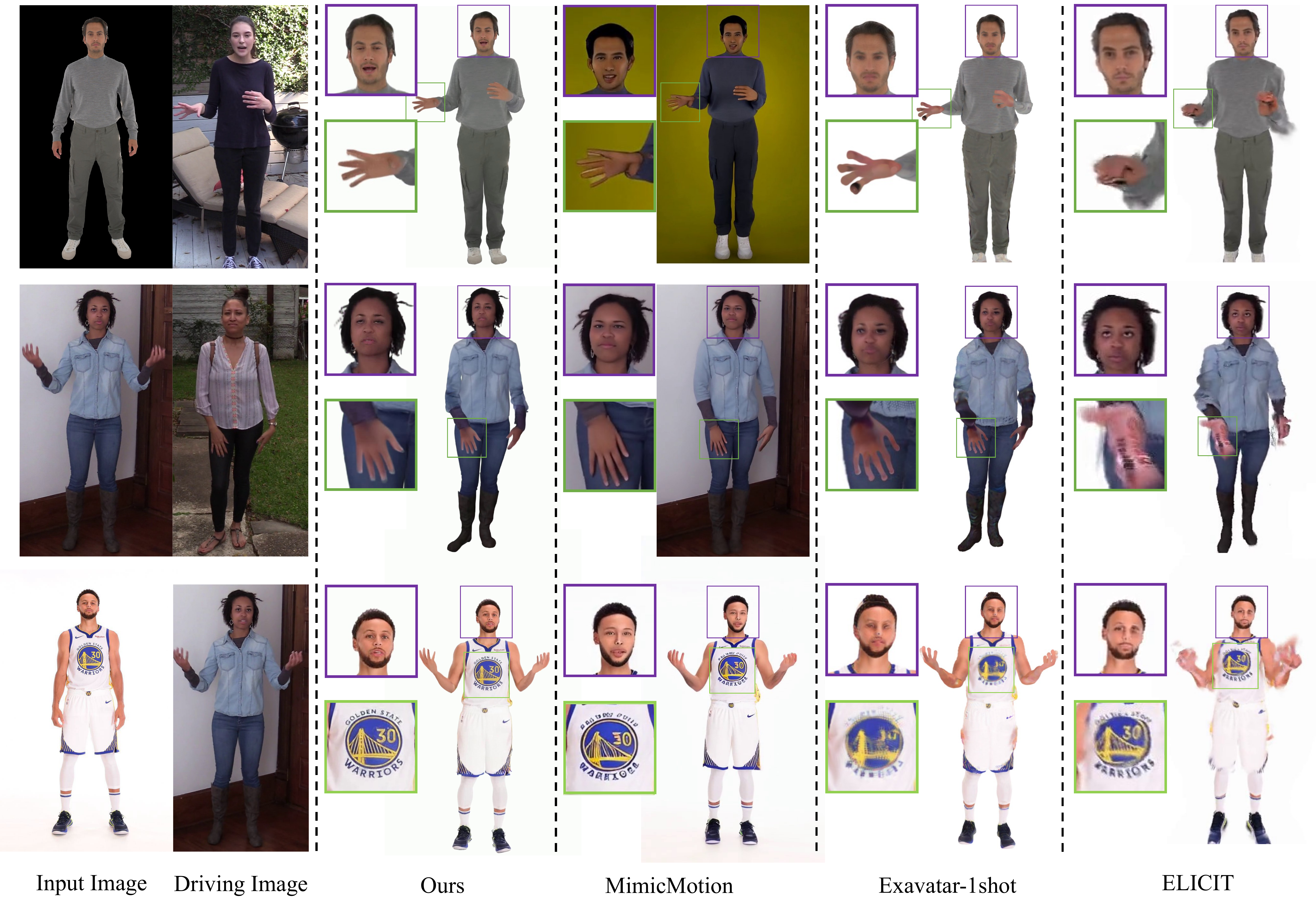}
  \caption{Qualitative comparisons with representative methods~\cite{mimicmotion2024,moon2024exavatar,huang2022elicit} in the cross-identity motion reenactment task. Our method achieves accurate and realistic animation with almost all fine details preserved and identity unchanged.}
  \vspace*{-5mm}
  \label{fig:driven}
\end{figure*}

Given a single image of the target person, we aim to reconstruct a 3D talking avatar that fully inherits the identity and enables natural animation. To address the challenge of complex dynamic modeling from imperfect pseudo videos, we adopt a tightly coupled 3DGS-mesh hybrid avatar representation (Sec.~\ref{sec:representation}). To generalize well to diverse gestures and facial movements, we generate imperfect video sequences of the target person driven by various motion sequences (Sec.~\ref{sec:pseudo_labels}). Finally, we introduce the carefully designed constraints and loss terms to train the representation from noisy videos effectively (Sec.~\ref{sec:constraints}). The entire pipeline is illustrated in Fig.~\ref{fig:pipeline}.

\subsection{Coupled 3DGS-Mesh Avatar}
\label{sec:representation}
Whole-body parametric mesh models facilitate human animation and provide good initialization, while 3DGS offers enhanced expressiveness and realistic rendering. To address the challenges of the one-shot task, we design a novel coupled 3DGS-mesh representation, which effectively integrates the geometric priors and surface regularization of the mesh without diminishing the expressive capability of the Gaussian field.

We couple the 3DGS field with the typical SMPL-X model, which is formulated as follows:
\begin{equation}
    M(\beta,\theta,\phi) = W(T(\beta,\theta,\phi),J(\beta),\theta,\mathcal{W}),
    \label{equ_smplx1}
\end{equation}
\begin{equation}
    T(\beta,\theta,\phi) = \bar{T} + B_S(\beta) + B_E(\phi) + B_P(\theta),
    \label{equ_smplx2}
\end{equation}
where $B_S(\cdot)$, $B_E(\cdot)$ and $B_P(\cdot)$ denote shape, expression, and pose blend functions respectively, with $\beta$, $\phi$ and $\theta$ representing the corresponding parameters. $\bar{T}$ is the template mesh, and $W(\cdot)$ is the standard LBS function that rotates the vertices in $T$ around the estimated joints $J$, smoothed by the blend weights $\mathcal{W}$. Since the shape code $\beta$ is fixed once registered, we will omit it in the later statements and denote $T=T(\beta,\theta,\phi)$ and $J=J(\beta)$ for simplicity. Inspired by~\cite{Pang_2024_CVPR, xiang2024flashavatar, abdal2024gaussian}, we initialize the 3D Gaussians on the canonical mesh surface using UV parameterization. For a 3D Gaussian located on a triangle $k=\{\mathbf{V_1},\mathbf{V_2},\mathbf{V_3}\}$ of mesh $T$ with barycentric coordinates $(u,v)$, the surface position is give by:
\begin{equation}
    Bary(T) = \mathcal{V}(k,u,v) = u\mathbf{V_1} + v\mathbf{V_2} + (1-u-v)\mathbf{V_3}.
    \label{equ_bary}
\end{equation}

\noindent
\textbf{Coupled 3DGS-Mesh Deformation.}
To model clothes, haircuts, and other complex regions that the SMPL-X mesh fails to handle, the deformation of 3D Gaussians is crucial. While previous works can achieve correct deformation with sufficient supervision, we need to impose additional constraints on the deformation field to prevent overfitting to the single input image and imperfect pseudo-labels, while also facilitating the integration of other modules.
Based on these considerations, we design two deformation fields in the canonical space and enforce their proximity: one represents the conventional Gaussian deformation, and the other represents the critical mesh deformation. This approach allows us to indirectly influence the Gaussian deformation by applying soft constraints on the deformed mesh. Moreover, since there is no strict binding between the deformed mesh surface and the Gaussians, complex regions are still effectively handled. The two deformation fields, together with the full-body animation and Gaussian Splatting process, are formulated as follows:
\begin{equation}
    G(\theta,\phi,\mathbf{P}) = SP(W(Bary(T)+dX,J,\theta,Bary(\mathcal{W})),\mathbf{P}).
\end{equation}
\begin{equation}
    M(\theta,\phi) = W(T+dT,J,\theta,\mathcal{W}),
\end{equation}
where $G(\cdot)$ represents the final rendered image, $SP(\cdot)$ and $\mathbf{P}$ denote the rendering process and the remaining properties of 3D Gaussians, and $dX$ and $dT$ represent the optimized deformations of the Gaussians and mesh vertices, respectively. We apply key regularizers on $dT$ and propagate soft constraints to $dX$ by encouraging $dX$ to align with $Bary(dT)$.


\subsection{Pseudo Labels Generation}
\label{sec:pseudo_labels}
In the one-shot image setting, many regions are unseen or occluded. To construct a complete avatar and ensure generalization to novel gestures and expressions, we turn our attention to recent advances in human motion diffusion models~\cite{mimicmotion2024,hu2024animate} and head animation techniques~\cite{deng2024,guo2024liveportrait}.

For leveraging diffusion-based generative models, the SDS loss~\cite{poole2022dreamfusion} is widely used in text/image-to-3D works~\cite{huang2024tech,zhang2024humanref,tang2023dreamgaussian,yang2024have}. Although diffusion models may introduce moderate deviations, the SDS technique remains effective in these studies. However, for tasks such as realistic dynamic avatar animation from a one-shot image, pose alignment is critical~\cite{moon2024exavatar,hu2024expressive}, and the misalignment introduced by 2D diffusion models cannot be overlooked. Therefore, directly applying SDS in our setting is suboptimal. Instead, we focus on extracting accurate information from pseudo frames synthesized by these generative models.

To generate whole-body pseudo labels, we collect a set of SMPL-X pose sequences $\{(\theta_{N_i},\phi_{N_i})_{i=1}^F\}_{j=1}^K$ from the TED Gesture Dataset~\cite{yoon2019robots} as input to the generative models. Since no unified generative model exists for both body and head simultaneously, we employ two separate approaches to generate pseudo labels for body gestures and facial expressions, respectively. Given a source image $I_S$ registered with the SMPL-X parameters $(\beta_S,\theta_S,\phi_S)$, we adopt MimicMotion~\cite{mimicmotion2024} with the collected pose sequences to generate various pseudo body frames: 
\begin{equation}
    I_N=Motion(I_S,D(\beta_S,\theta_N,\phi_N)), 
\end{equation}
where $D(\cdot)$ denotes the mapping from SMPL-X parameters to DWPose~\cite{yang2023effective}. We randomly set the root body pose in $\theta_N$ for each segment to increase viewpoint generalization, with pitch $\in (-30^\circ, 30^\circ)$ and yaw $\in (-10^\circ, 10^\circ)$. After that, we re-track $I_N$ to obtain more accurate pose parameters $(\hat{\theta}_N, \hat{\phi}_N)$, while keeping the shape code $\beta_S$ fixed. The re-tracking process is crucial, as it overcomes the misalignment introduced by 2D diffusion models and helps achieve precise and realistic results. For the head region, we adopt Portrait4D-v2~\cite{deng2024} to generate various pseudo frames of the target person performing diverse expressions, also driven by the videos from the TED Gesture Dataset.

\subsection{Objective Functions}
\label{sec:constraints}
Assisted by our novel hybrid 3DGS-mesh avatar representation, we introduce several carefully designed regularization terms to stabilize the avatar reconstruction process and effectively extract the correct information from both the one-shot input and the imperfect pseudo-labels.

\noindent
\textbf{Mesh-related Constraints.}
We use the following loss functions to apply soft constraints on the Gaussian field based on the mesh:
\begin{equation}
    \begin{aligned}
        \mathcal{L}_{SC} =& \lambda_{\textrm{normal}}\mathcal{L}_{\textrm{normal}}+\lambda_{\textrm{M}}\mathcal{L}_{\textrm{M}}\\
        +& \lambda_{\textrm{MGC}}\mathcal{L}_{\textrm{MGC}}+ \lambda_{\textrm{lap}}\mathcal{L}_{\textrm{lap}}.
    \end{aligned}
\end{equation}
Here, $\mathcal{L}_{\textrm{normal}}$ is the normal consistency loss applied to the deformed mesh surface, ensuring surface normal consistency post-deformation. $\mathcal{L}_{M}$ is the mask loss, which measures the discrepancy between the ground truth mask and the mask of the deformed mesh rendered via~\cite{Laine2020diffrast}. These two losses work together to regulate the behavior of $dT$, while influencing Gaussian deformation $dX$ through the mesh-Gaussian consistency loss: 
\begin{equation}
    \mathcal{L}_{\textrm{MGC}}=\|dX-Bary(dT)\|_1. 
\end{equation}
Additionally, we compute the Laplacian smoothing loss $\mathcal{L}_{\textrm{lap}}$ for $dX$ along with the scaling and RGBs of the 3D Gaussians, based on their initial connectivity state on the canonical mesh surface.

\noindent
\textbf{Perceptual Supervision of Pseudo-Labels.}
The synthesized pseudo-labels exhibit noticeable artifacts, such as image distortion and lack of 3D consistency, especially in the results generated with MimicMotion. These artifacts cannot be fully corrected through pose alignment alone. As a result, pixel-aligned losses like L1 or MSE often lead to issues like texture flickering, blurring, and identity changes. However, the high-level human structure is consistently well-preserved in the source image, which aids human recognition and is empirically referred to as perceptual similarity. Previous works~\cite{zhang2018unreasonable,mechrez2018contextual} have used deep neural networks to capture these perceptual structure features. We adopt the LPIPS perceptual loss~\cite{zhang2018unreasonable}, using VGG~\cite{simonyan2014very} as the backbone. By learning deep perceptual human features from dynamic poses and preserving detailed information from the source image, our method is able to conduct realistic and complete talking avatar animations. Thus, for the pseudo frames, we primarily use the following perceptual loss:
\begin{equation}
     \mathcal{L}_{\textrm{diff}} = \lambda_{\textrm{LPIPS}}\mathcal{L}_{\textrm{LPIPS}}(I_N, G(\hat{\theta}_N,\hat{\phi}_N,\mathbf{P})).
\end{equation}

\noindent
\textbf{Per-pixel Supervision of the Input Image.}
For the source image $I_S$ with pose $(\theta_S,\phi_S)$, we use the regular L1 loss, SSIM loss, and mask loss between $I_S$ and the splatted image $G(\theta_S,\phi_S,\mathbf{P})$:
\begin{equation}
     \mathcal{L}_{S} = \lambda_{\textrm{RGB}}\mathcal{L}_{\textrm{RGB}}+\lambda_{\textrm{SSIM}}\mathcal{L}_{\textrm{SSIM}} + \lambda_{\textrm{alpha}}\mathcal{L}_{\textrm{alpha}}.
\end{equation}
The final loss function is the sum of all the aforementioned losses.

\begin{figure*}
  \centering
  \includegraphics[width=1.0\linewidth]{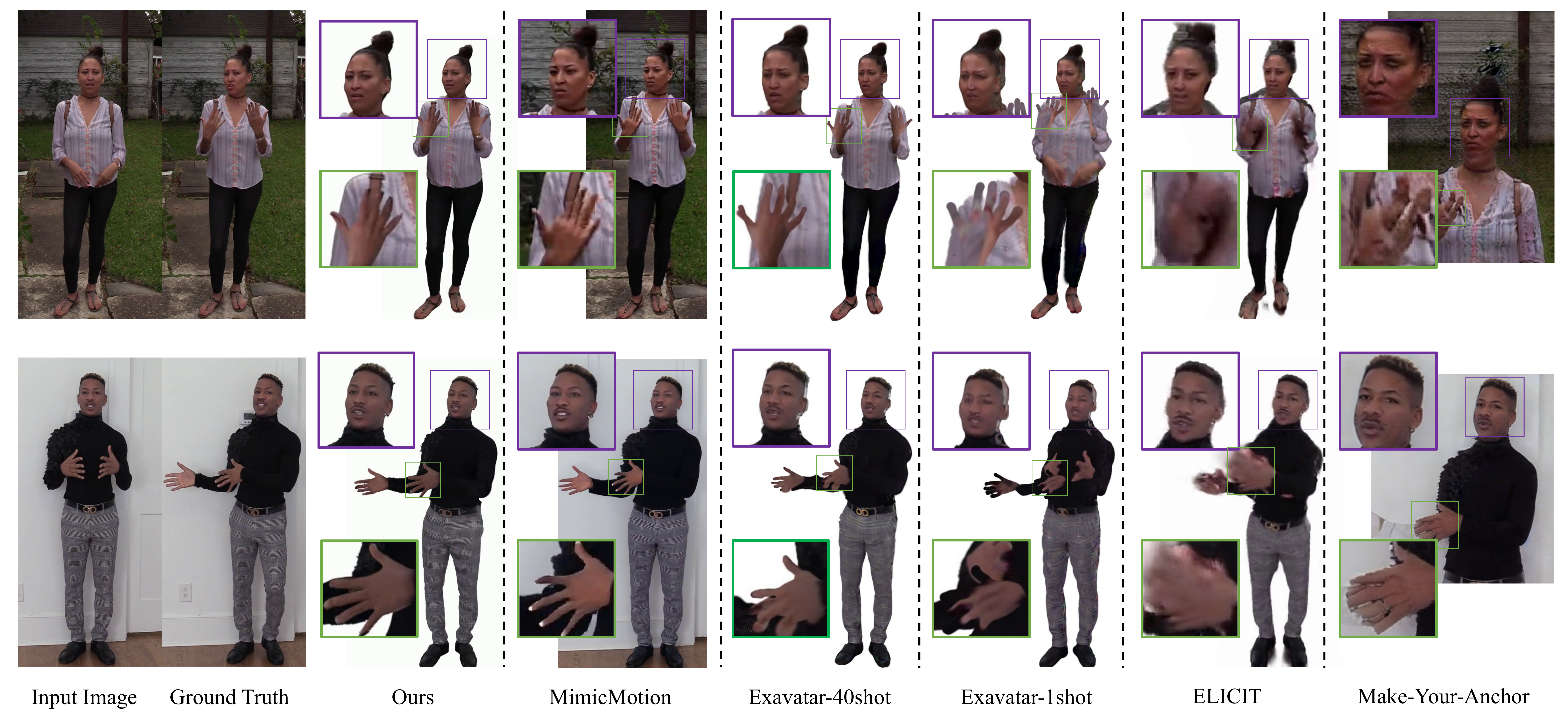}
  \caption{Qualitative comparisons with representative methods~\cite{mimicmotion2024,moon2024exavatar,huang2022elicit,huang2024make} in the self-driven motion reenactment task. Our method well models facial and hand regions, which match the input image most in global identity preservation and local details modeling, even compared with some methods trained on captured videos.}
  \vspace*{-5mm}
  \label{fig:selfdriven}
\end{figure*}

\subsection{Implementation Details}
\noindent
\textbf{Gaussian Field.}
Similar to~\cite{hu2024gaussianavatar}, we adopt an isotropic Gaussian field for better generalization to novel viewpoints and poses. Specifically, we fix the opacity $\alpha=1$, the rotation quaternion $q = (1, 0, 0, 0)$, and only use a scalar scaling factor $\hat{s}$ for the Gaussians. We set the UV map resolution to 512 and use approximately 150,000 Gaussians in total.

\noindent
\textbf{Optimization.}
We train our models using the Adam optimizer~\cite{kingma2014adam} with $\beta = (0.9, 0.999)$. The learning rates for the Gaussian parameters are the same as those in the official implementation, while the learning rates for both the Gaussian and mesh deformation fields are set to $\eta = 1e^{-4}$. For the loss weights, we set $\lambda_{\textrm{normal}}$, $\lambda_{\textrm{M}}$, $\lambda_{\textrm{MGC}}$, $\lambda_{\textrm{lap}}$, $\lambda_{\textrm{LPIPS}}$, $\lambda_{\textrm{RGB}}$, $\lambda_{\textrm{SSIM}}$, $\lambda_{\textrm{alpha}}$ to $1e^{-2}$, $1e^{-1}$, $1e^{1}$, $1e^{2}$, $2e^{-1}$, $8e^{-1}$, $1e^{-1}$, and $4e^{-1}$, respectively. We start adding the perceptual loss at the 2000th step, while the other losses are used throughout the entire training process. We empirically find that longer training results in better quality and does not lead to training collapse.
\section{Experiments}
\label{sec:exper}

\subsection{Dataset} 
We use the poses and expressions processed from 100 videos of the TED Gesture Dataset~\cite{yoon2019robots} as the motion sequences during training. For evaluation, the one-shot input and driving poses are primarily sourced from the ActorsHQ~\cite{icsik2023humanrf} and the Casual Conversations Dataset~\cite{hazirbas2021towards}. All videos and images are cropped and resized to a fixed aspect ratio of 9:16, with videos sampled at 30 FPS. For foreground segmentation, we use BiRefNet~\cite{zheng2024birefnet}, and for human pose tracking, we employ the custom fitting procedure from ExAvatar~\cite{moon2024exavatar}.

\subsection{Comparison with Representative Methods}

We compare our method with several representative works, including: (1) ExAvatar~\cite{moon2024exavatar}, a recent SOTA work that models human avatars with a mesh-based 3D Gaussian field. (2) ELICIT~\cite{huang2022elicit}, a one-shot NeRF-based animatable avatar. (3) MimicMotion~\cite{mimicmotion2024}, a general 2D pose-guided human video diffusion model. (4) Make-Your-Anchor~\cite{huang2024make}, a 2D diffusion-based method that is fine-tuned on a one-minute video clip of the individual to enhance identity information.

Note that for ExAvatar, since it takes short videos as input, we compare with two versions of it: ExAvatar-40shot, which uses 40-shot images, and ExAvatar-1shot, which uses one-shot images as input. For Make-Your-Anchor, as we find it does not perform well on one-shot input, we only compare it with available video input by fine-tuning it on a short video clip.

\noindent \textbf{Qualitative Comparisons.} \cref{fig:driven} and ~\cref{fig:selfdriven} present a qualitative comparison between our method and the other representative approaches. For \cref{fig:driven}, we use the pose of a different identity as the driving signal. For \cref{fig:selfdriven}, we compare the performance of these methods on the self-driven pose reenactment task, using subjects that have corresponding video data.

As observed, ExAvatar-1shot tends to overfit the input image and fails to recover accurate textures in regions occluded by hands. Even with 40-shot input, ExAvatar still struggles with incomplete knowledge and fails to handle novel gestures effectively. MimicMotion generates relatively reasonable results but is constrained by its training distribution and struggles with identity consistency across frames, often leading to appearance mismatches and identity changes. ELICIT, which uses a NeRF-based representation, ensures 3D consistency but relies on SMPL for geometry, which neglects the hand region. This coarse semantic proxy fails to support complex hand reconstruction or animation. Make-Your-Anchor, although pre-trained on multiple identities to learn human motion priors, requires a long video with sufficient movement and appearance data to adapt to new identities. It struggles with short fine-tuning videos and fails to recover fine details, especially for gestures outside its fine-tuning distribution.

\begin{figure}
  \centering
  \includegraphics[width=1.0\linewidth]{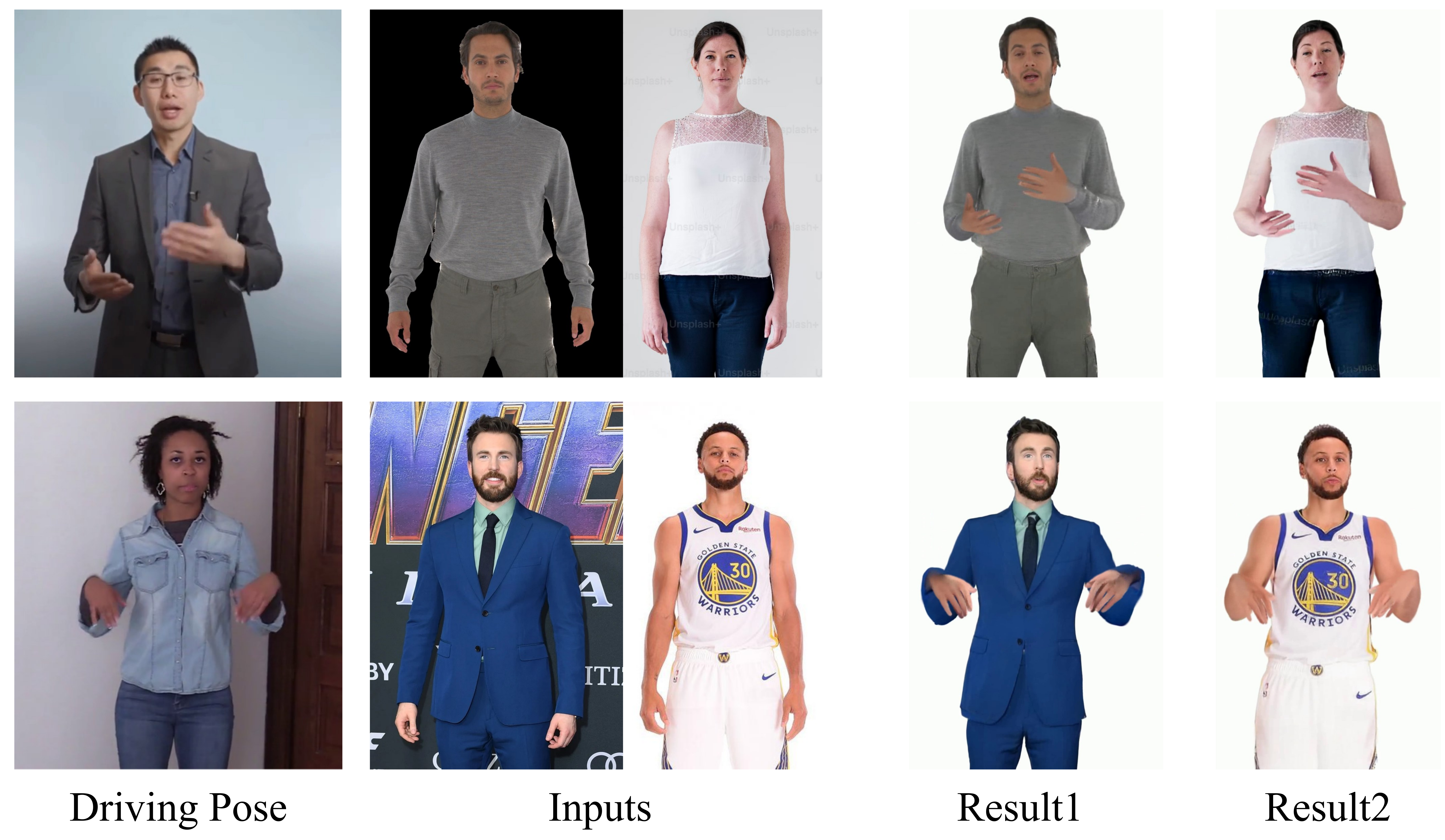}
  \caption{More examples of cross-identity pose reenactment. Different subjects can be accurately animated with the same poses.}
  \vspace*{-2mm}
  \label{fig:samepose}
\end{figure}

In contrast, our method generates animatable and expressive talking avatars from a single input image. Our results preserve fine human details and achieve natural animation with excellent rendering quality. We provide additional examples of cross-identity pose reenactment in~\cref{fig:samepose}. Using the same driving pose, identities with completely different attributes can be driven in the same way, thanks to the SMPL-X model and the 3DGS-mesh coupled representation.

\noindent \textbf{Quantitative Comparison.} \cref{tab:ssim} presents a quantitative comparison between our model and other methods. Although the one-shot image animation task typically lacks a strict test set for quantitative comparison, we perform the evaluation on the self-driven task and use five common metrics for comparison: Mean Squared Error (MSE), L1 distance, PSNR, SSIM, and LPIPS. Our method outperforms all others across these metrics, demonstrating superior realism and 3D consistency in the results. However, we believe that these metrics do not fully capture the quality or capabilities of the methods, especially for the one-shot image animation task. We encourage readers to refer to the video results of our method for a more comprehensive and objective evaluation.

\begin{table}
  \centering
  \resizebox{\columnwidth}{!}{
    \begin{tabular}{@{}lcccc@{}}
      \toprule
      Metrics & ELICIT & ExAvatar & MimicMotion & Ours \\
      \midrule
      MSE($10^{-3}$)$\downarrow$ & 5.65 & 3.93 & 2.69 & \textbf{1.22} \\
      L1($10^{-2}$)$\downarrow$ & 1.65 & 1.24 & 1.41 & \textbf{0.84} \\
      PSNR$\uparrow$ & 22.56 & 24.22 & 25.84 & \textbf{29.31} \\
      SSIM($10^{-1}$)$\uparrow$ & 9.21 & 9.24 & 9.18 & \textbf{9.43} \\
      LPIPS($10^{-2}$)$\downarrow$ & 6.60 & 4.09 & 3.89 & \textbf{2.99} \\
      \bottomrule
    \end{tabular}
  }
  \caption{Quantitative comparisons with representative methods~\cite{mimicmotion2024,moon2024exavatar,huang2022elicit} on self-driven data. Our method outperforms others in pixel-wise error metrics, realism evaluation metrics and perceptual quality metrics. (ExAvatar here denotes ExAvatar-40shot.)}
  \vspace*{-3mm}
  \label{tab:ssim}
\end{table}

\begin{figure*}
  \centering
  \includegraphics[width=1.0\linewidth]{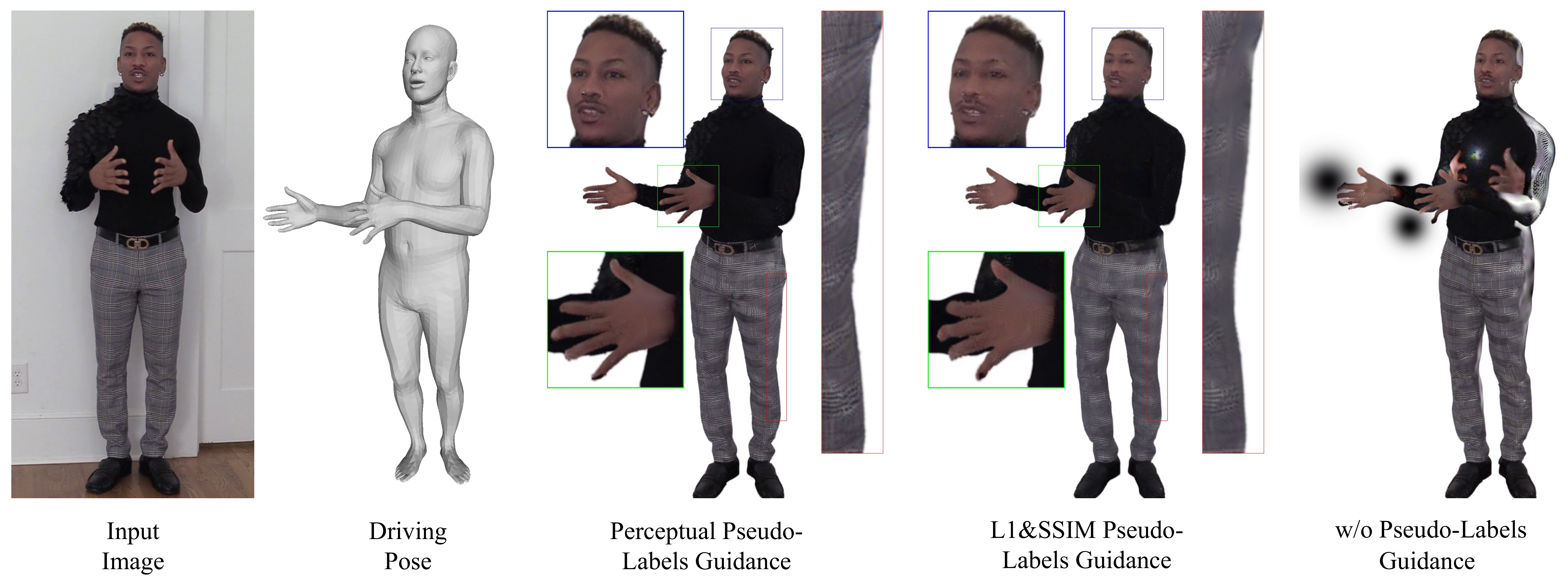}
  \caption{Perceptual diffusion guidance is of great importance to inpainting unseen regions and modeling natural and realistic textures.}
  \vspace*{-5mm}
  \label{fig:aba-diff}
\end{figure*}

\subsection{Ablation Studies}
We conduct ablations studies on several critical components of the proposed method.

\begin{figure}
  \centering
  \includegraphics[width=1.0\linewidth]{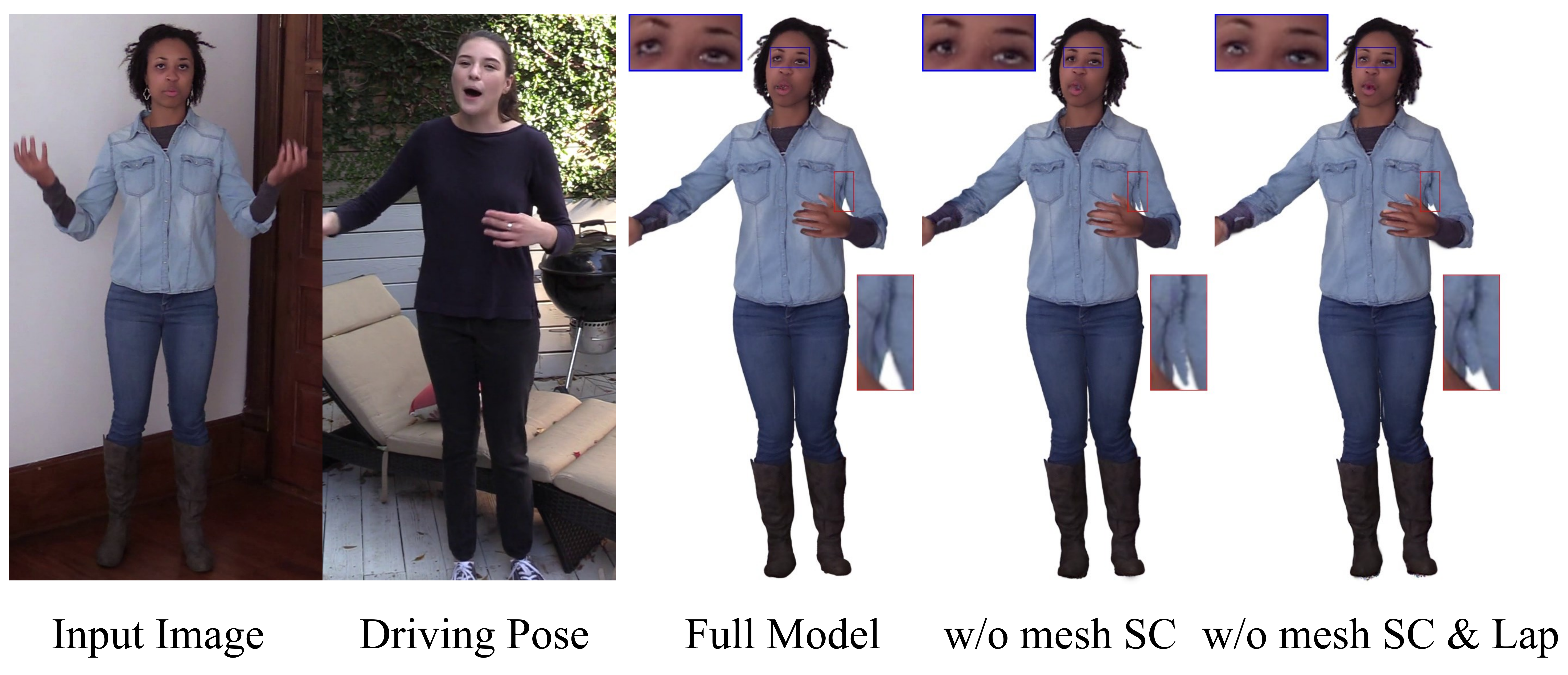}
  \caption{Soft mesh constraints together with Gaussian Laplacian help preserve geometric integrity and model fine details.}
  \label{fig:aba-sc}
\end{figure}

\noindent
\textbf{Mesh-related Constraints.} Our designed hybrid 3DGS-mesh avatar representation and the corresponding regularizations significantly enhance the expressiveness and integrity of the final results, as demonstrated in~\cref{fig:aba-sc}. When soft constraints for mesh deformation are omitted, geometric artifacts appear in specific regions. Similarly, without the Gaussian Laplacian loss, the Gaussian field fails to capture fine details, further compromising the overall quality.

\begin{figure}
  \centering
  \includegraphics[width=1.0\linewidth]{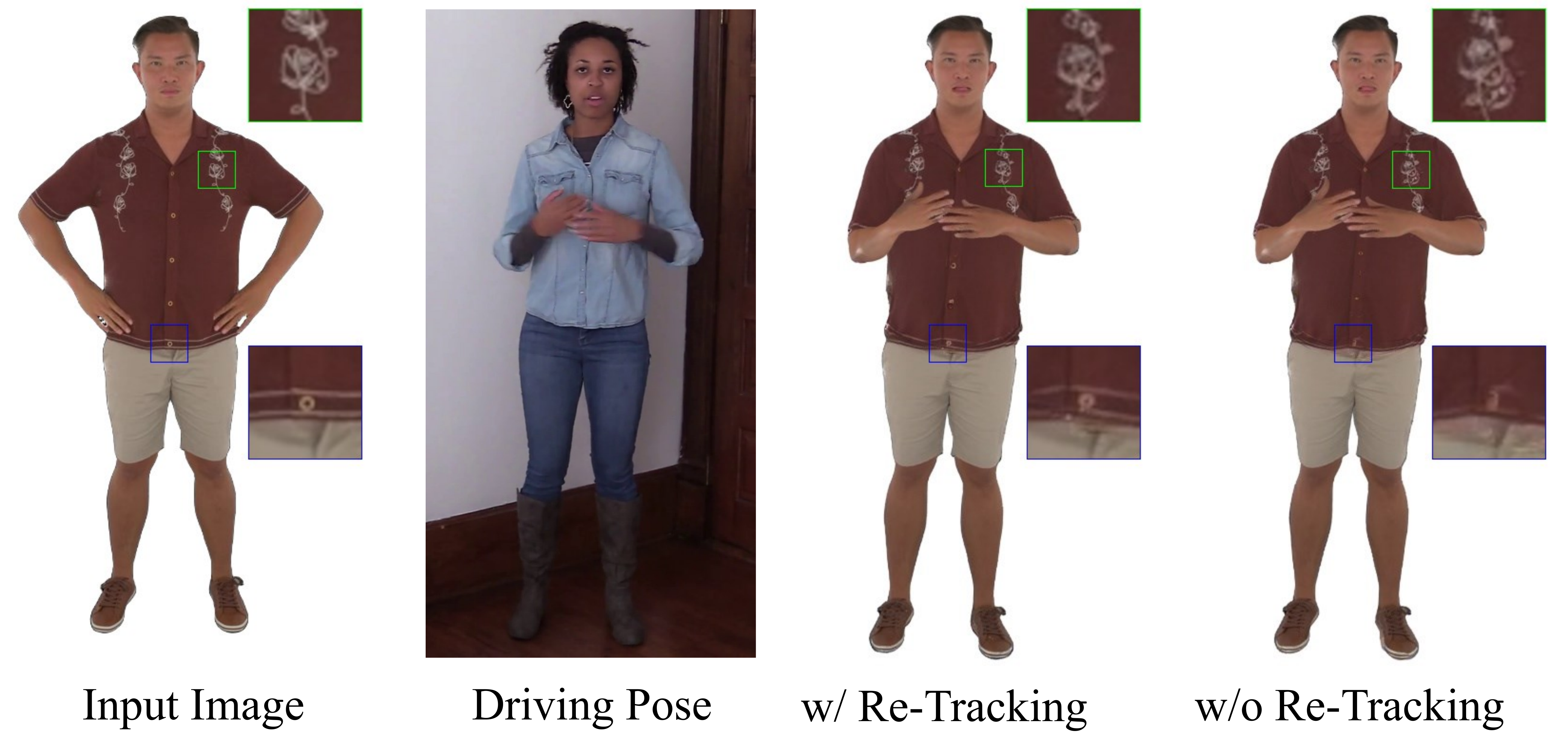}
  \caption{Re-Tracking step preserves better texture structures and avoid texture loss.}
  \vspace*{-5mm}
  \label{fig:aba-retrack}
\end{figure}

\noindent
\textbf{Perceptual-based Pseudo-Labels Guidance.} Pseudo-labels guidance is the core concept behind this work, enabling the challenging task of generating animatable and expressive talking avatars from a single input image. As shown in~\cref{fig:aba-diff}, without diffusion guidance, we are unable to inpaint unseen regions effectively, and the result suffers from severe overfitting to the input image. Additionally, due to the misalignment and inconsistency of the pseudo-output generated by motion diffusion models, using pixel-based losses like L1 for diffusion guidance results in overly smooth outputs and struggles to capture fine details, especially in facial and hand regions. In contrast, the perceptual diffusion guidance we employ not only preserves the full personalized attributes of the input image but also inpaints unseen regions with more natural and consistent textures.

\noindent
\textbf{Re-Tracking Step.} The re-tracking procedure applied to the pseudo-output of motion diffusion models helps mitigate the misalignment introduced by 2D diffusion models, preventing texture errors and detail loss. As shown in~\cref{fig:aba-retrack}, the re-tracking step successfully recovers more accurate texture structures.

\section{Conclusion}
\label{sec:conclu}

In this paper, we introduce a novel pipeline for creating expressive talking avatars from a single image. We propose a coupled 3DGS-Mesh avatar representation, incorporating several key constraints and a carefully designed hybrid learning framework that combines information from both the input image and pseudo frames. Experimental results demonstrate that our method outperforms existing techniques, with our one-shot avatar even surpassing state-of-the-art methods that require video input. Considering its simplicity in construction and ability to generate vivid, realistic animations, our method shows significant potential for practical applications of talking avatars across various fields.

\noindent\textbf{Limitations.} The approach relies on accurate registration between the input image and the parametric human mesh, and severe mismatches, especially in regions like fingers, can cause optimization issues and result in incorrect textures. Additionally, rendering large views or extending to full 360° human reconstruction remains difficult due to current limitations in human motion diffusion models and the lack of data for large, novel viewpoints. Future work will explore integrating semantic information from large language models and static priors from 3D reconstruction to address these limitations.

{
    \small
    \bibliographystyle{ieeenat_fullname}
    \bibliography{main}
}


\clearpage



\twocolumn[{%
    \renewcommand\twocolumn[1][]{#1}%
        \maketitlesupplementary
        \appendix
        \includegraphics[page=1,width=\textwidth]{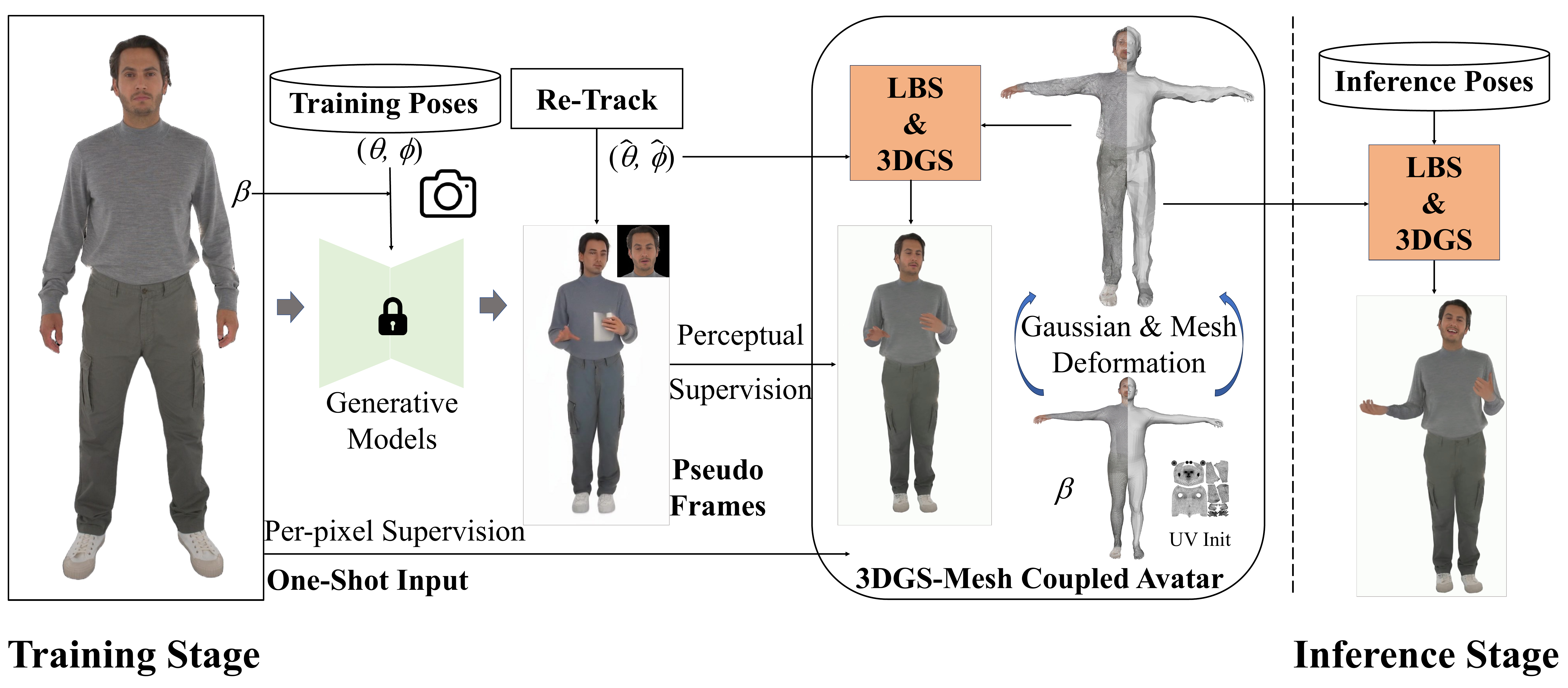}
        \vspace{-10mm}
        \captionof{figure}{A detailed illustration of our pipeline.}
        \label{fig:detailpipe}
    \vspace{5mm}
}]

\section{Detailed Pipeline}
For a comprehensive understanding, we present a detailed illustration of our pipeline in Fig.~\ref{fig:detailpipe}.

\section{Limitation}
\begin{figure}
  \centering
  \includegraphics[width=1.0\linewidth]{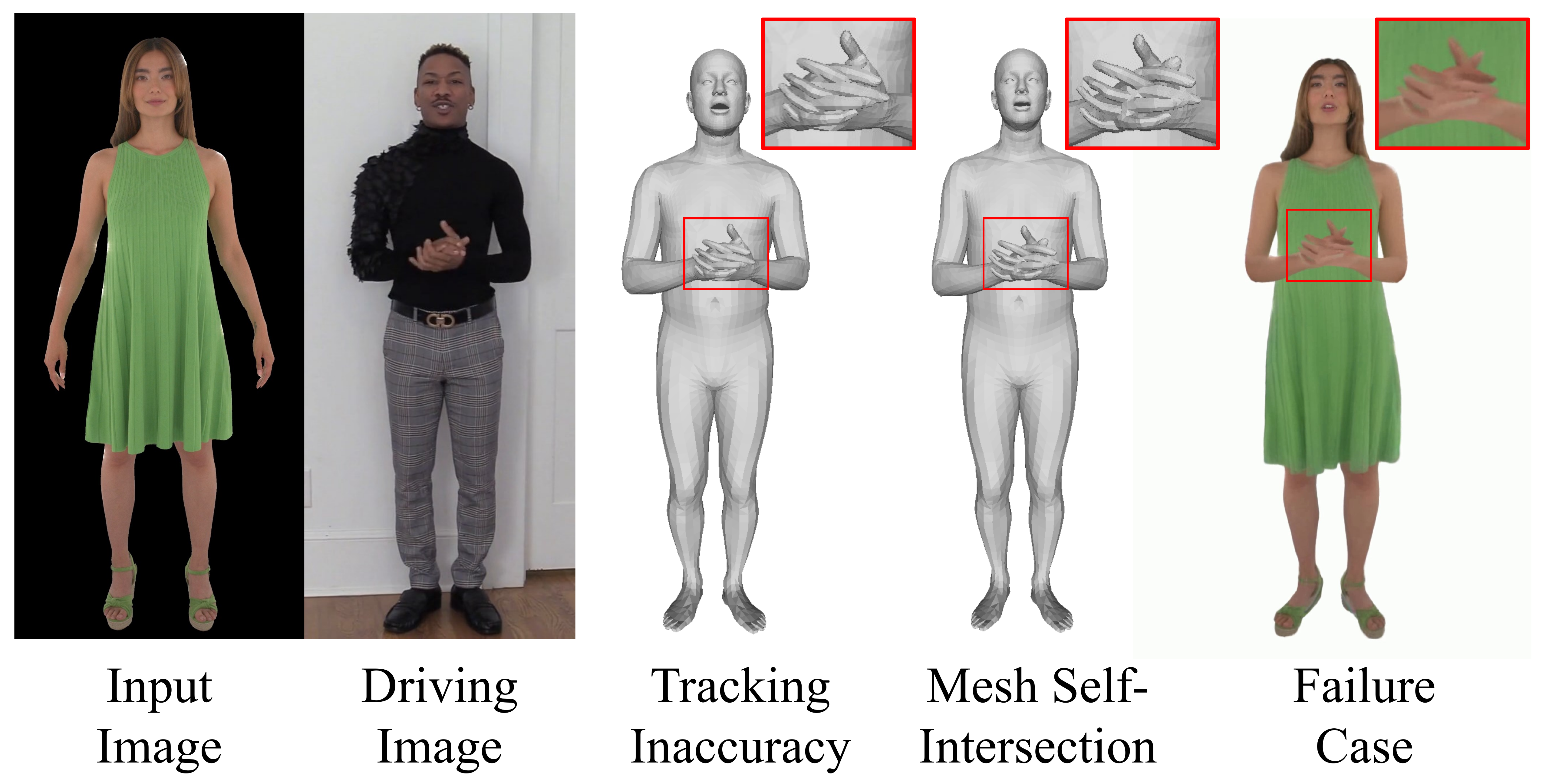}
  \vspace*{-5mm}
  \caption{Inaccurate tracking and finger self-intersection during cross-identity animation.}
  \label{fig:limit}
\end{figure}

\begin{figure}
  \centering
  \includegraphics[width=1.0\linewidth]{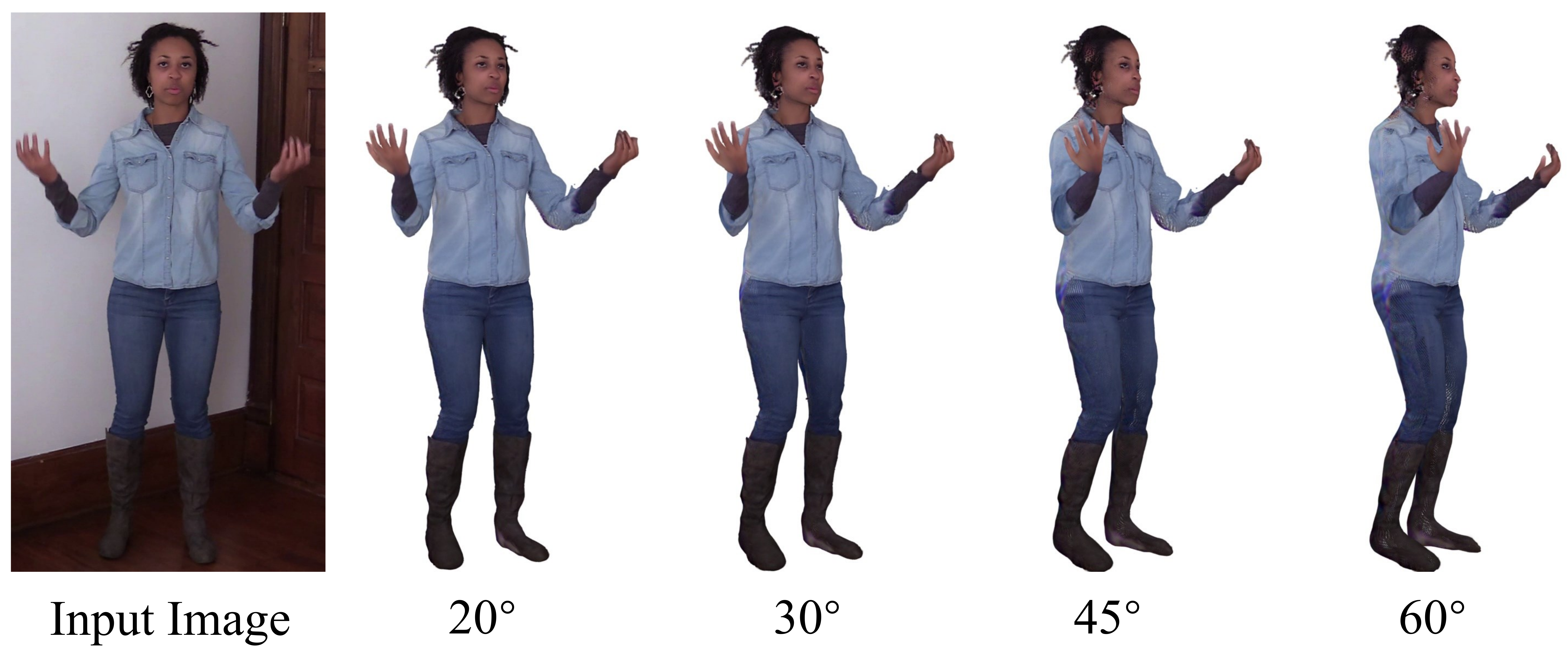}
  \vspace*{-5mm}
  \caption{Novel view results across diverse angles.}
  \vspace*{-5mm}
  \label{fig:nvs}
\end{figure}

\noindent
\textbf{Tracking and Large-View Rendering.} Accurate SMPL-X tracking is essential for mesh-based avatar representation. Our method relies on precise registration between the input image or video and the parametric human mesh, which can be compromised by tracking inaccuracies. Additionally, self-intersection may occur when we conduct cross-identity animation, particularly observed in the finger area, as illustrated in~\cref{fig:limit}. Furthermore, our method demonstrates robust performance within a view range of -30 to 30 degrees. However, its performance degrades for larger viewing angles, as demonstrated in~\cref{fig:nvs}.

\section{Broader Impact}
\label{sec:impact}
Our work enables the reconstruction of expressive whole-body talking avatars from a single photo, allowing for realistic animations with vivid body gestures and natural expression changes. We consider this a significant advancement in the research and practical applications of multimodal digital humans. However, this technology carries the risk of misuse, such as generating fake videos of individuals to spread false information or harm reputations. We strongly condemn such unethical applications. While it may not be possible to entirely prevent malicious use, we believe that conducting research in an open and transparent manner can help raise public awareness of potential risks. Additionally, we hope our work can inspire further advancements in forgery detection technologies.

\end{document}